\documentclass[runningheads]{llncs}
\usepackage[T1]{fontenc}
\usepackage{multirow}
\usepackage{graphicx}
\usepackage{color}
\usepackage{stix}
\usepackage[Export]{adjustbox}
\usepackage{tabularx}
\usepackage{amsmath}

\usepackage{algorithm}
\usepackage{algpseudocode}
\algrenewcommand\algorithmicrequire{\textbf{Input:}}
\algrenewcommand\algorithmicensure{\textbf{Output:}}
\usepackage{xcolor}
\usepackage{multirow}
\usepackage{color, colortbl,booktabs}
\definecolor{Gray}{gray}{0.9}
\definecolor{myGreen}{rgb}{0.0, 0.6, 0.0}
\usepackage{arydshln}
\usepackage{caption}
\usepackage{subcaption}
\usepackage{layouts}
\usepackage{hyperref}
\usepackage{bbding}
\usepackage[
    style=ieee,
    citestyle=numeric-comp,
    backend=biber,
    hyperref=true,
    maxnames=1,
    maxbibnames =2,
    minnames=1,
    mincrossrefs=1,
    natbib=true,
    sorting=none,
    sortcites=true,
    url=false,
    doi=false,
    eprint=false
]{biblatex}
\AtEveryBibitem{%
  \clearname{translator}%
  \clearlist{publisher}%
  \clearfield{pagetotal}%
\clearfield{pagetotal}%
\clearfield{issn}%
\clearfield{editor}%
\clearfield{month}%
\clearfield{notes}%
\clearfield{language}%
}
\DeclareFieldFormat
  [article,inbook,incollection,inproceedings,patent,thesis,unpublished]
  {title}{{#1}}

\begin{document}

\title{CLAIM: \textbf{C}linically-Guided \textbf{L}GE \textbf{A}ugmentation for Real\textbf{i}stic and Diverse \textbf{M}yocardial Scar Synthesis and Segmentation}
\titlerunning{CLAIM}

\author{Farheen	Ramzan\inst{1,2}\thanks{Corresponding Author: framzan1@sheffield.ac.uk}, Yusuf Kiberu\inst{3}, Nikesh Jathanna\inst{3,4}, Shahnaz	Jamil‑Copley\inst{3,4}, Richard H. Clayton\inst{1,2\star\star}, Chen (Cherise) Chen\inst{1,2}\thanks{Co-Senior Authorship}}

\authorrunning{F. Ramzan et al.}

\institute{Insigneo Institute for \emph{in-silico}, Medicine, University of Sheffield, Sheffield, UK. \and School of Computer Science, University of Sheffield, Sheffield, UK.\and Department of Cardiology, Nottingham University Hospitals NHS Trust, Nottingham, UK.\and School of Medicine, University of Nottingham, Nottingham, UK.}
\maketitle

\begin{abstract}
Deep learning-based myocardial scar segmentation from late gadolinium enhancement (LGE) cardiac MRI  has shown great potential for accurate and timely diagnosis and treatment planning for structural cardiac diseases. However, the limited availability and variability of LGE images with high-quality scar labels restrict the development of robust segmentation models. To address this, we introduce CLAIM: \textbf{C}linically-Guided \textbf{L}GE \textbf{A}ugmentation for Real\textbf{i}stic and Diverse \textbf{M}yocardial Scar Synthesis and Segmentation framework, a framework for anatomically grounded scar generation and segmentation. At its core is the SMILE module (Scar Mask generation guided by cLinical knowledgE), which conditions a diffusion-based generator on the clinically adopted AHA 17-segment model to synthesize images with anatomically consistent and spatially diverse scar patterns. In addition, CLAIM employs a joint training strategy in which the scar segmentation network is optimized alongside the generator, aiming to enhance both the realism of synthesized scars and the accuracy of the scar segmentation performance. Experimental results show that CLAIM produces anatomically coherent scar patterns and achieves higher Dice similarity with real scar distributions compared to baseline models. Our approach enables controllable and realistic myocardial scar synthesis and has demonstrated utility for downstream medical imaging task. Code is available at  \href{https://github.com/farheenjabeen/CLAIM-Scar-Synthesis}{https://github.com/farheenjabeen/CLAIM-Scar-Synthesis}

\keywords{Diffusion model \and Image synthesis \and Cardiac magnetic resonance imaging \and Myocardial infarction. \and Segmentation.}
\end{abstract}

\section{Introduction}
Late Gadolinium Enhancement (LGE) imaging is used to assess myocardial viability by highlighting affected areas with increased contrast uptake allowing non-invasive diagnosis of myocardial infarction (MI), ischemic and non-ischemic cardiomyopathies as well as other cardiac pathologies \cite{jenista2023revisiting, kiberu2024myocardial}. However, the automated detection of hyperenhanced areas (referred to as scar) is a challenging task due to poorly defined contours owing to low soft tissue contrast, along with imaging contrast differences and spatial and temporal variations across different scanners, sequences and study protocols. Moreover, the heterogeneous nature of myocardial scar across patients such as structural irregularity, and diversity in spatial location, shape and scale of pathological target, especially in the presence of image noise and artifacts, adds more to the difficulty.

Although deep learning (DL) based methods have achieved impressive performance to automate general medical segmentation tasks, the development of automated robust and generalizable models for precise scar detection and segmentation \cite{lalande2022deep, li2023myops} remains challenging. DL-based methods require large annotated, high-quality datasets for model training, which is expensive and extremely difficult to acquire for scar segmentation tasks. Finding precise contours of scar from LGE images is hard and requires substantial clinical expertise. In addition, security and privacy concerns limit dataset sharing across multiple centers \cite{qayyum2020secure}. These challenges have significantly impeded the development of scalable automated solutions for scar segmentation in the absence of large scale datasets.

To address this dilemma, a promising approach is to generate synthetic data to supplement real-world data and enhance the training of DL models ~\cite{savage2023synthetic}. Recently, pathology synthesis has garnered significant interest from the research community, where generative models have played important roles to generate images with diverse pathology ~\cite{chen2024towards, dorjsembe2024conditional}, achieving promising results for various types of medical imaging data such as tumor synthesis in light sheet microscopy \cite{horvath2022metgan}, colon polyps in colonoscopy \cite{pishva2023repolyp, du2023arsdm}, lung nodules in computed tomography images \cite{han2019synthesizing, jin2021free}, brain tumors in MRI \cite{billot2023synthseg} and tumors in various organs \cite{wu2025freetumor}. Generative models learn from pathological data and can generate synthetic data containing pathology or lesions with increased variety, which in turn can be used for downstream applications.

Yet, there are still some unsolved challenges in generative methods. A major challenge is data hallucination where unintended alterations and contextual artifacts appear in background (non-lesion) regions. These methods may also fail to generate lesions with sufficient diversity in size, location, and texture, which can lead to suboptimal down-stream task (e.g. segmentation) performance. To alleviate these problems, recent studies have explored the use of conditional diffusion models to synthesize pathological images from normal cases \cite{chen2024towards, zhang2024lefusion, tian2025lesiondiffusion, chou2025ultrasound}, which can increase data diversity while better controlling the background regions to remain unaffected.

These methods condition on both normal images and pathology masks as input and synthesize images by using an inpainting scheme \cite{lugmayr2022repaint} to fill the pathological regions with appropriate texture. Rather than using the background as conditional inputs \cite{rombach2022high}, these methods explicitly integrate forward-diffused background directly during the iterative diffusion process to better preserve the background regions. Another key component for achieving \emph{diverse} and controllable pathology synthesis is introducing diverse, yet realistic pathology masks for conditioning so that the generated synthetic dataset can present a wide range of pathological variations while remaining anatomically plausible. However, most mask generation algorithms mainly rely on the statistics of existing training datasets, which can still be biased ~\cite{zhang2024lefusion}. Also, there is still no guarantee that the synthesized foreground regions are semantically meaningful.

In this work, to address the scarcity of annotated LGE scar datasets required for robust and generalizable deep learning-based scar segmentation, we also resort to advance conditional diffusion models for synthesizing high-quality, diverse datasets. Different from prior work, we focus on generating clinically plausible pathology masks that take clinically meaningful segments of the left ventricular (LV) myocardium into account. This enables the synthesis of images with anatomically consistent and spatially diverse scar patterns within the LV myocardium. To this end, we propose CLAIM (\textbf{C}linically-Guided \textbf{L}GE \textbf{A}ugmentation for Real\textbf{i}stic and Diverse \textbf{M}yocardial Scar Synthesis and Segmentation) framework for myocardial scar generation and segmentation. At its core is the SMILE module (Scar Mask generation guided by cLinical knowledgE), which conditions a diffusion-based generator on the clinically adopted AHA 17-segment model ~\cite{Cerqueira2002-AHA} for left ventricular myocardium to synthesize realistic and diverse LGE images. In addition, CLAIM employs a joint training paradigm in which the scar segmentation network is optimized alongside the generator, aiming to enhance both the realism of synthesized scars and the accuracy of the scar segmentation performance. For training and evaluation, we used pathological and normal cases from the EMIDEC dataset \cite{lalande2020emidec}. We also validated the performance of our model on a private out-of-domain dataset \cite{jathanna2023nottingham, shahnazprivatedata} for robustness evaluation. We further investigated the effect of the synthesized images on the performance of nnUNet \cite{isensee2021nnu} based segmentation models for the LGE scar segmentation task. Our main contributions are summarized as follows:
\begin{itemize}
    \item We present CLAIM which introduces a joint modelling paradigm to jointly train a conditional diffusion model guided by clinical knowledge for LGE image synthesis with diverse scar patterns and a robust cardiac scar segmentation model, enjoying the mutual benefits with both improved scar synthesis and segmentation performance.
    \item  To the best of our knowledge, this is the first work to incorporate the American Heart Association (AHA) 17-segment model into the image generative process of an DL-based diffusion framework for myocardial scar synthesis. By leveraging this standardized anatomical representation, CLAIM ensures that synthetic scar patterns are not only spatially diverse but also clinically interpretable.
    \item  We demonstrate that CLAIM improves segmentation performance by augmenting training data with clinically realistic and diverse scar patterns on synthetic LGE images via both qualitative and quantitative evaluation. We validate the approach using both the EMIDEC dataset (including pathological and normal cases).
    \item We further assess the robustness and generalizability of CLAIM using an independent out-of-domain dataset, showing its improved robustness against domain shift.
\end{itemize}

\section{CLAIM: \textbf{C}linically-Guided \textbf{L}GE \textbf{A}ugmentation for Real\textbf{i}stic and Diverse \textbf{M}yocardial Scar Synthesis and Segmentation}
We present CLAIM, a novel framework that is capable of producing high-quality, diverse synthetic pathological images to augment the training of deep scar segmentation model, with improved accuracy and robustness. At a high level, the framework consists of a conditional diffusion model based on Denoising Diffusion Probabilistic Model (DDPM) for scar synthesis and a scar segmentation model, which is trained in an alternating fashion to iteratively improve both synthesis quality and segmentation performance.
\begin{figure}[t]
\centering
\includegraphics[keepaspectratio=true, width=0.9\textwidth]{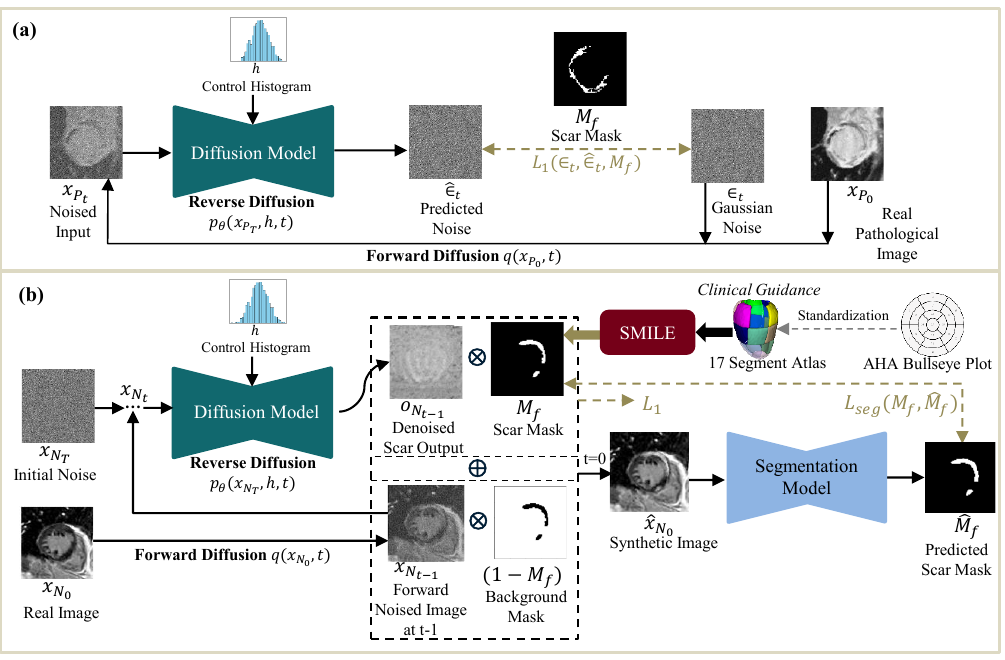}
\caption{Overview of CLAIM. (a) illustrates the training of a diffusion model with real pathological images, while (b) presents detailed pathological image generation process with diversified scar patterns, thanks to the clinical knowledge guided mask generation module (SMILE) (Sec.~\ref{section2.2}), the generated synthetic images are then used to finetune a scar segmentation model to enhance the quality of scar synthesis via joint training.} \label{fig1}
\end{figure}

\subsection{Joint Modelling for Enhanced Scar Synthesis and Segmentation}
\label{section2.1}
Our diffusion model is adapted from the one used in the LeFusion~\cite{zhang2024lefusion}, a lesion-focused diffusion model, consisting of a forward diffusion process $q$ that adds noise to images and a reverse diffusion process $p_{\theta}$ that recovers images from the noisy input, conditioned on the specified scar regions. But different from LeFusion, we introduce a scar segmentation model to inform the training of diffusion model and enhance the quality of pathological image synthesis.

As shown in Fig.~\ref{fig1}, given a pathological image with scar presented, we perform the forward diffusion process $q (x_{P_0},t$) \cite{ho2020denoising, rombach2022high} to get a noisy input ${x}_{P_{t}}$ to the diffusion model at time step t, $t\in [1,T]$. This will add a time-dependent Gaussian noise $\epsilon_t$ to an original pathological image (forward diffusion). The diffusion model  (often implemented as a UNet \cite{ronneberger2015u}, parameterized by $\theta$) is then trained to produce the predicted  noise $\hat\epsilon_t$ at time step $t$ (reverse diffusion $p_\theta$), with the guidance of a control histogram $h$ that produced from the training dataset, which controls the texture of lesion area, as originally proposed in Lefusion~\cite{zhang2024lefusion}. A scar-focused noise prediction loss  $L_1(\epsilon, \hat\epsilon; M_{f}$) is applied to ensure that the model focuses only on manipulating the specified scar regions $M_f$:
\begin{equation}
  L_1(\epsilon_t, \hat{\epsilon_t}; M_{f}) = \| M_{f}*\epsilon_t - M_{f}*\hat{\epsilon_t} \|^2_{2}].    \label{eq:2}
\end{equation}

Different from LeFusion, we introduce a scar segmentation model for the joint training of the diffusion model, which has been pretrained on real pathological cases\footnote{The segmentation model is based on the most popular and state-of-the-art medical segmentation backbone: nnUNet \cite{isensee2021nnu}.} to ensure the quality of synthetic images. In order to jointly train both models, we perform online data augmentation to generate synthetic images using the diffusion model,which can be used to finetune the segmentation model. During online data generation, the diffusion model takes an LGE image from a subject $x_{N_0}$ and a scar mask $M_f$ as input to generate a pathological image $\hat{x}_{N_0}$. Of note, the input data can be healthy subject or pathological cases where the scar mask can thus be randomly generated or extrapolated based on the existing scar mask for data augmentation (See Sec.~\ref{section2.2}). The same $L_1$ loss is applied for the training of the diffusion model. The synthesized image is then passed to the segmentation network to predict the scar $\hat{M}_f$. A segmentation loss $L_{seg}$ computed on the synthetic data is applied to enhance the training of the segmentation network:
\begin{equation}
    L_{seg}(M_f, \hat{M}_f) = L_{Dice}(M_f, \hat{M}_f) + L_{WCE}(M_f, \hat{M}_f)
    \label{eq:3}.
\end{equation}
Here $L_{Dice}$ is a soft Dice loss and $L_{WCE}$ is a weighted cross entropy loss to account for the severe class imbalance problem presented in this task.

During the synthetic image generation process, in order to synthesize pathological images with diverse scar patterns with background preserving, we start by sampling a noise vector $x_{N_{T}} \sim \mathcal{N}(0,1)$ and then iteratively refine the scar region through the backward process using the diffusion model from $t$ to $t-1$, which is defined as: $
    \hat{x}_{N_{t-1}} = o_{N_{t-1}} \odot M_f + {x}_{N_{t-1}} \odot (1-M_f)$~\cite{zhang2024lefusion},
where $o_{N_{t-1}} \sim p_{\theta}$ is the synthesized scar predicted through the reverse diffusion process $p_{\theta}$;  ${x}_{N_{t-1}}$is a forward noised input at $t-1$ derived from the forward diffusion process $q$ by adding randomly sampled noise $\epsilon_t$ to the real image ${x}_{N_{0}}$. The  background mask $(1-M_f)$ is applied to $x_{N_{t-1}}$ to preserve the non-lesion regions. 

\subsection{SMILE: \textbf{S}car \textbf{M}asks Incorporating c\textbf{L}inical knowledg\textbf{E}}\label{section2.2}
\begin{figure}[t]
\centering
\includegraphics[keepaspectratio=true, width=0.8\textwidth]{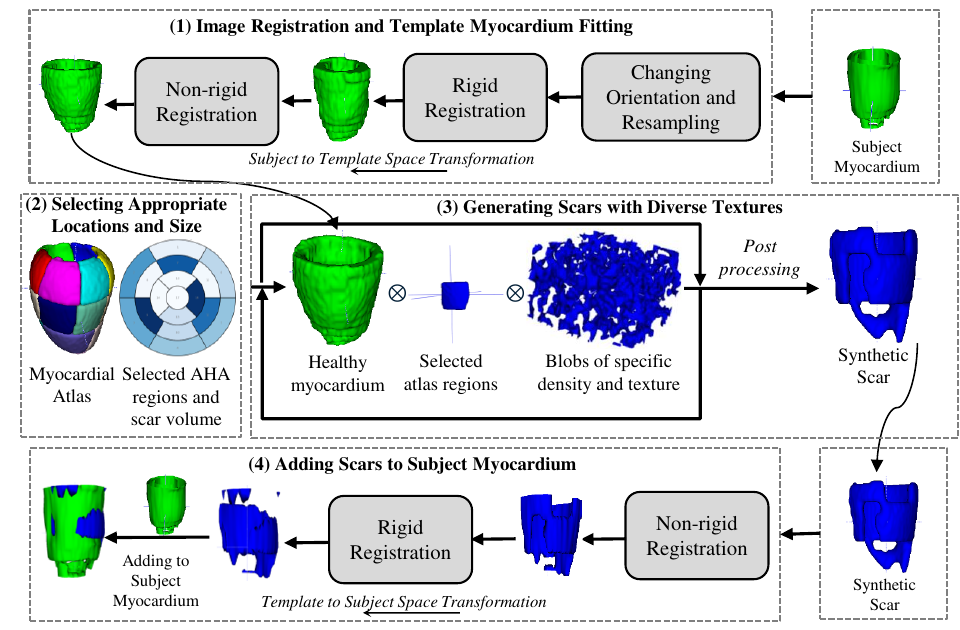}
\caption{Overview of SMILE (Scar Masks Incorporating cLinical knowledgE)} \label{fig2}
\end{figure}
To further enhance the realism and diversity of scar masks, we introduce SMILE: \textbf{S}car \textbf{M}asks Incorporating c\textbf{L}inical knowledg\textbf{E}, which is able to control the location and shape of the masks in the myocardium guided by an atlas. Specifically, SMILE utilizes standardized LV myocardial segmentation and nomenclature for cardiac imaging defined by the AHA \cite{selvadurai2018definition, Cerqueira2002-AHA}. This standard divides the LV myocardium into 17 segments in basal, middle and apical regions, which enables the mapping of myocardial segments to coronary arterial topography and is important for localizing, reporting, evaluating, and assessing abnormalities in disease management, both in clinical practice and research settings. We are the first to incorporate AHA in the AI model development process. The steps in SMILE are visualized in Fig ~\ref{fig2} and detailed as following:\\
\textbf{1. Image Registration and Myocardium Fitting:} In order to generate plausible scar masks that lie within the boundary of the myocardium and with reasonable shape, we propose to sample scar masks according to the AHA-17 definition. To identify the AHA segment for diverse subjects, firstly, we register each subject's myocardium masks (usually available in public scar datasets) to a template space with AHA-17 segments\footnote{The AHA-17 atlas/template is provided by Bai W., et al. \cite{bai2015bi}.}. During this procedure, orientation correction, re-sampling and rigid registration followed by non-rigid registration (i.e. fast symmetric forces demons method \cite{tang2016effective}) are applied.\\
\textbf{2. Selecting Appropriate Locations and Size for Scar Synthesis:} In the AHA scheme, the myocardium is divided into basal (regions 1-6), middle (regions 7-12) and apical (regions 13-17) slices. In order to select diverse and unique scar locations, our method starts by randomly selecting one or more myocardial regions from basal, middle and apical slices. We used adjacency maps to ensure that the selected regions were adjacent to each other. To ensure diverse scar sizes, we used scar volume per myocardial region as a metric. We sampled a list of scar volumes for each myocardial region from a uniform distribution.\\ \textbf{3. Generating Scars with Diverse Textures:} 
After obtaining the list of regions and corresponding scar volumes, we generated textured scar masks. For each selected AHA region and scar volume, a 3D blob matching the template size was created. Texture was controlled via anisotropy, and volume by adjusting porosity, followed by erosion with a randomly selected kernel (1$\times$1 to 7$\times$7). Each blob was placed into the corresponding AHA region and then combined with the myocardium to ensure anatomical plausibility. This process was repeated across selected regions, and the resulting scar regions were merged into a complete scar mask. Post-processing included hole filling, Gaussian smoothing, and removal of isolated regions.\\
\textbf{4. Adding Scars to Subject Myocardium: }Finally, we transformed the scar masks from template to subject space using non-rigid (fast symmetric forces demons) and then rigid registration (based on mutual information) and add it to the original scar masks (if applicable).

\section{Experiments and Results}
\subsection{Datasets}
\textit{Public Dataset:} The EMIDEC dataset \cite{lalande2020emidec} provides 100 labeled short-axis DE-MRI (Delayed Enhancement MRI). This includes 67 pathological and 33 normal cases. The segmentation labels contain annotations for left ventricle (LV), healthy myocardium (MYO), myocardial infarction (MI) and microvascular obstruction (PMO). In this study we only focus on MI.\\
\textit{Private Dataset:} We used 80 pathological cases from a private dataset provided by our clinical partners to serve as out-of-domain data for robustness evaluation purpose. This dataset is provided by the Nottingham University Hospitals \cite{jathanna2023nottingham, shahnazprivatedata}, which consists of short axis LGE-MRI scans with scar labelled by our cardiologists. Detailed dataset information can be found in Table~\ref{tab1:dataset}.

\subsection{Experimental Setup}
\begin{table}[t]
\centering
\caption{Characteristics of the LGE-MRI datasets used in this study}
\label{tab1:dataset}
\resizebox{0.8\textwidth}{!}{
\begin{tabular}{|c|cc|c|c|c|c|}
\hline
\multirow{2}{*}{Dataset} & \multicolumn{2}{c|}{Subjects} & \multirow{2}{*}{\begin{tabular}[c]{@{}c@{}}Pixel Spacing\\ (mm)\end{tabular}} & \multirow{2}{*}{\begin{tabular}[c]{@{}c@{}}Slice Thickness\\ (mm)\end{tabular}} & \multirow{2}{*}{\begin{tabular}[c]{@{}c@{}}Slice Distance\\ (mm)\end{tabular}} & \multirow{2}{*}{Scar Volume (mean$\pm$std)} \\ \cline{2-3}
 & \multicolumn{1}{c|}{Normal} & Pathological &  &  &  &  \\ \hline
EMIDEC \cite{lalande2020emidec} & \multicolumn{1}{c|}{33} & 67 & 1.25$\times$1.25, 2$\times$2 & 8 & 10 & 23.68$\pm$15.81 \\ \hline
Private \cite{shahnazprivatedata, jathanna2023nottingham} & \multicolumn{1}{c|}{0} & 80 & 0.89$\times$0.89,1.7$\times$1.7 & 8-10 & 10 & 14.33$\pm$13.62 \\ \hline
\end{tabular}}
\end{table}
We split 67 pathological cases from EMIDEC dataset into 57 real pathological cases for training image synthesis models and 10 for testing the segmentation models. We split 33 normal cases into 28 cases for testing image synthesis models and 5 for validation during joint training. We used synthetic images to train segmentation models with different real and synthetic subset settings for downstream scar prediction task. For evaluating the segmentation models, we used 10 pathological cases from EMIDEC dataset (in-domain data) and 80 pathological cases from private dataset (out-of-domain data).

For image synthesis, the images were cropped from the center based on the bounding boxes obtained from the corresponding masks. Then the image intensity was rescaled between -1 and 1. Random flipping based data augmentation was applied for training image synthesis models. For segmentation task, images were normalized by applying z-score normalization and random data augmentation was applied on the fly by using various transforms \cite{isensee2021nnu}. This data augmentation scheme was used for training all the segmentation models. Both CLAIM and CLAIM (J) were trained for 50,000 epochs with a learning rate of 1e-4 using Adam optimizer (consistent with LeFusion \cite{zhang2024lefusion}). All the segmentation models were trained for 1000 epochs and stochastic gradient descent (SGD) method was used as an optimizer with an initial learning rate of 1e-2 (reduced continuously using the 'polyLR' scheme: $(1-epoch/epoch_{max})^{0.9}$).\\

\subsection{Method for Comparison}
For scar mask generation, we compared our results with DiffMask used in LeFusion\cite{zhang2024lefusion}. DiffMask employs a control sphere and a boundary mask followed by a smoothing kernel to diversify the size and location of the lesion masks. Such a design helps the diffusion model to capture varying shape correlations and spatial distributions. For image synthesis, we compared our results with LeFusion \cite{zhang2024lefusion}. \textit{CLAIM Variants (Ours):} There are two variants of our method. To investigate the effect of using a pretrained segmentation model on the quality of the synthesized images, we trained CLAIM without joint training, where only diffusion-based generator is trained. CLAIM (J) denotes joint training of both models. \textit{Baseline:} We trained a segmentation model for scar prediction task, to serve as a baseline, by using only real pathological cases from EMIDEC without any synthetic data from generative models.

\subsection{Visual Quality Assessment on Image Synthesis}
\subsubsection{Effectiveness of SMILE with Clinical Knowledge}
\begin{figure}[t]
\centering
\includegraphics[keepaspectratio=true, scale=0.8]{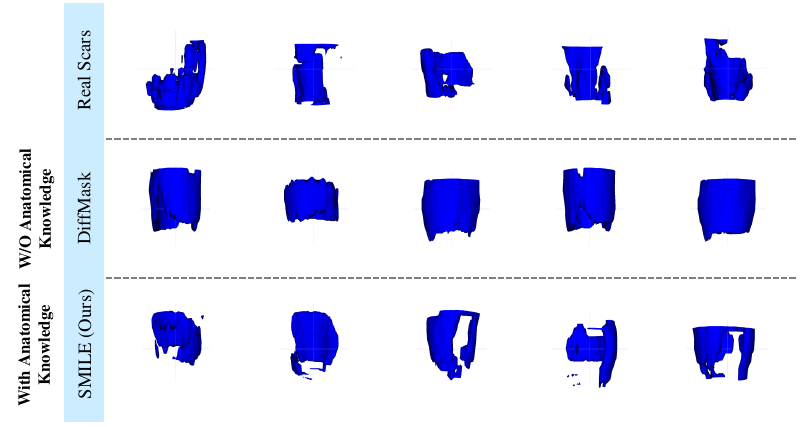}
\caption{Visualization of real and synthetic masks generated by DiffMask \cite{zhang2024lefusion} and SMILE.} \label{fig3}
\end{figure}

We visualized real scar masks and compared them with the synthetic masks generated by a mask generation algorithm called DiffMask  used in LeFusion~\cite{zhang2024lefusion} and SMILE (ours) as shown in Fig. ~\ref{fig3}. Our method is capable of generating scars of diverse shape patterns, locations and texture and it is evident  that the distribution of our scar masks are closer to the  real masks.

Different from our method, DiffMask does not consider clinical knowledge, which uses a simple ball shaped control sphere along with bounding mask to control the location/size and boundary of the generated masks. This ball shaped sphere can be used for enclosed organs (such as lungs, kidneys etc.). However, for a complex anatomical structure such as the left ventricle of the heart, it is unable to capture the characteristic of myocardial scars due to the hollow cylinder-like shape of the left ventricular wall, which encloses the scars. In addition, DiffMask relies on the statistics of existing training datasets, which can be biased. DiffMask generated masks are much bigger and sometimes can even cover the entire myocardium. Similar findings can be also found in their own plotted results (Figure A2) on the myocardium scar generation task\cite{zhang2024lefusion}. By contrast, our generated masks display varied forms, sizes and locations.

\subsubsection{Effectiveness of Joint Training}
\begin{figure}[t]
\centering
\includegraphics[keepaspectratio=true, scale=0.8]{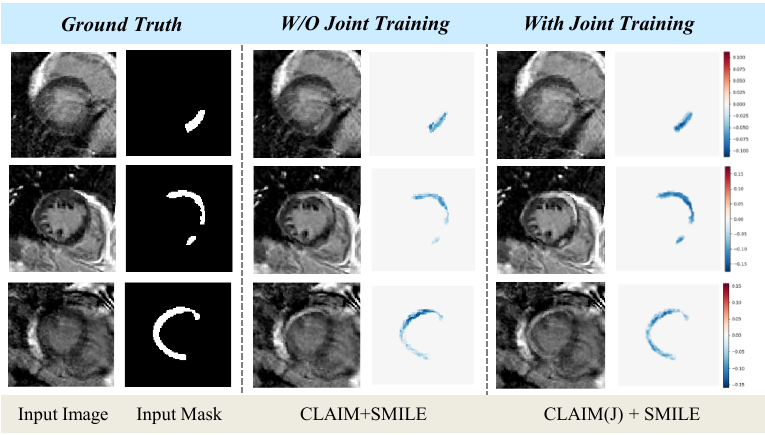}
\caption{Visualization of synthetic pathological images generated by different methods given normal images and scar masks. The difference between the input normal image and the synthetic pathological image is shown beside each image.} \label{fig4}
\end{figure}

To assess the effect of joint training paradigm on pathological image generation, we also visualized the generated images by ablating the joint training mechanism in Fig. ~\ref{fig4}. We also plotted the difference image between input normal images and output synthetic images on the scar regions  where higher values represent brighter scar regions (indicating the presence of scars in the synthetic image). It can be seen that the background regions remain preserved in all methods. However, we can see that each method captures intensity variations of scar at varying levels. We observe that the synthetic images generated by model trained with joint training shows high intensity (brighter) in the scar regions as compared to the models trained without the joint training. It is more evident when smaller scar regions are given, CLAIM with joint training can better model the abnormal tissues. 

\subsection{Effectiveness of CLAIM for Scar Segmentation}
To validate the utility of CLAIM for the downstream segmentation task, we trained different segmentation models for the scar segmentation task under different training settings. The backbones are the same nnUNet~\cite{isensee2021nnu}. We used different subsets of synthetic images along with real pathological images to train nnUNet \cite{isensee2021nnu} for scar segmentation task and compared it to the state-of-the-art method: LeFusion~\cite{zhang2024lefusion} with the original proposed mask generation algorithm (DiffMask) and with our proposed mask generation algorithm (SMILE). We also compared the segmentation model trained with (CLAIM (J)) and without the joint training of the diffusion model (CLAIM). For fair comparison, we re-implemented and repeated all the experiments.

\noindent\textbf{In-Domain Test:}
We evaluated the performance of different models on scar segmentation task by using 10 real pathological cases from the EMIDEC dataset (in-domain data). The results are given in Table~\ref{tab2}. We report the performance of each method using Dice score, precision, sensitivity and specificity. We compared the results of CLAIM and CLAIM (J) with LeFusion \cite{zhang2024lefusion} as well as with the baseline model (the naive nnUNet without any augmented data from generative models). We observed that our method based on a joint training paradigm achieved superior Dice scores and precision for all groups of segmentation models. As the size of the training data increases with synthetic data for the segmentation task, the downstream segmentation performance in terms of Dice score and precision also improves, indicating the effectiveness of synthetic data generated via CLAIM. 
\begin{table}[t]
\centering
\caption{Scar segmentation results on EMIDEC dataset \cite{lalande2020emidec}. P: 57 real pathological cases. N’: (28$\times$2) synthetic pathological cases generated from 28 normal cases. N”: (28$\times$5) synthetic pathological cases generated from 28 normal cases. P’: (57$\times$1) synthetic pathological cases generated from 57 real pathological cases. CLAIM: without joint training, CLAIM (J): with joint training.}
\label{tab2}
\scalebox{0.8}{
\begin{tabular}{|c|c|c|c|c|c|c|}
\hline
\textbf{Methods} & \textbf{Scar Mask} & \textbf{Dataset} & \textbf{Dice Score} & \textbf{Precision} & \textbf{Sensitivity} & \textbf{Specificity} \\ \hline
Baseline & --- & P & 58.89 & 70.99 & 62.35 & 99.35 \\ \hline
LeFusion \cite{zhang2024lefusion} & DiffMask & \multirow{4}{*}{P+N’} & 59.50 & 68.48 & 62.68 & 99.02 \\ \cline{1-2} \cline{4-7} 
LeFusion & \multirow{3}{*}{SMILE} &  & 59.33 & 69.71 & \textbf{63.97} & 98.98 \\ \cline{1-1} \cline{4-7} 
CLAIM (Ours) &  &  & 62.04 & 73.28 & 63.65 & 99.03 \\ \cline{1-1} \cline{4-7} 
CLAIM (J)  (Ours) &  &  &\textbf{ 63.07} & \textbf{74.13} & 63.52 & \textbf{99.11} \\ \hline
LeFusion \cite{zhang2024lefusion} & DiffMask\cite{zhang2024lefusion} & \multirow{4}{*}{P+N’’} & 60.43 & 68.96 & 63.40 & 99.04 \\ \cline{1-2} \cline{4-7} 
LeFusion & \multirow{3}{*}{SMILE} &  & 60.09 & 70.22 & 64.34 & 99.03 \\ \cline{1-1} \cline{4-7} 
CLAIM (Ours) &  &  & 62.98 & 73.73 & \textbf{63.51} & \textbf{99.08} \\ \cline{1-1} \cline{4-7} 
CLAIM (J)  (Ours) &  &  & \textbf{63.27} & \textbf{73.77} & 64.32 & 99.07 \\ \hline
LeFusion \cite{zhang2024lefusion} & DiffMask\cite{zhang2024lefusion} & \multirow{4}{*}{P+P'+N’’} & 60.91 & 71.95 & 62.34 & 99.04 \\ \cline{1-2} \cline{4-7} 
LeFusion & \multirow{3}{*}{SMILE} &  & 62.63 & 71.14 & \textbf{65.69} & 98.99 \\ \cline{1-1} \cline{4-7} 
CLAIM (Ours) &  &  & 62.87 & 73.61 & 64.05 & 98.99 \\ \cline{1-1} \cline{4-7} 
CLAIM (J)  (Ours) &  &  & \textbf{63.53} & \textbf{73.88} & 64.71 & \textbf{ 99.07} \\ \hline
\end{tabular}}
\end{table}

\begin{figure}[!tbh]
\vspace*{-3mm}
\centering
\begin{subfigure}{.5\textwidth}
  \centering
  \includegraphics[keepaspectratio=true, scale=0.4]{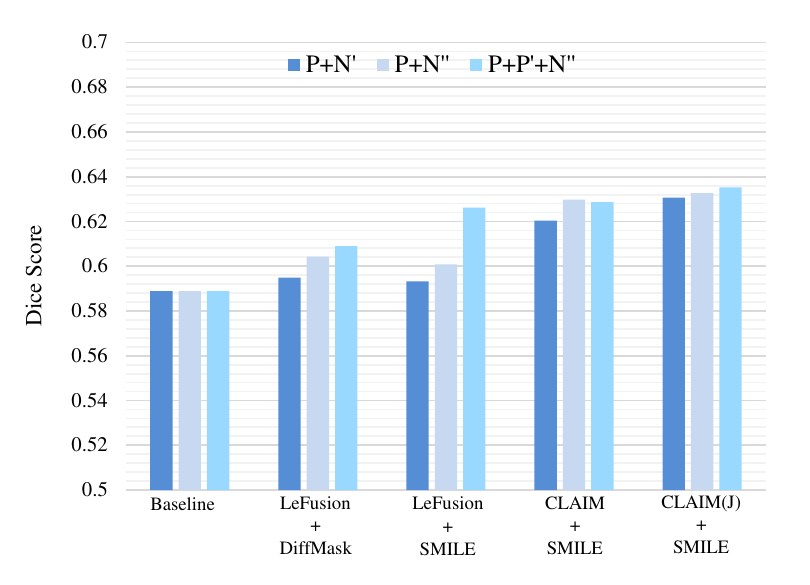}
  \caption{EMIDEC (in-domain data)}
  \label{fig:sub5.1}
\end{subfigure}%
\begin{subfigure}{.5\textwidth}
  \centering
\includegraphics[keepaspectratio=true, scale=0.4]{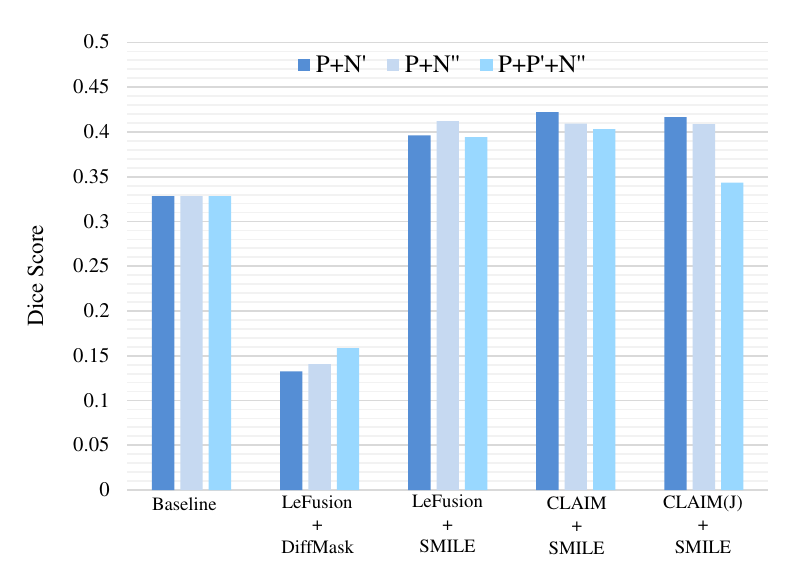}
  \caption{Private (out-of-domain data)}
  \label{fig:sub5.2}
\end{subfigure}
\caption{Comparing performance of different methods for scar segmentation on EMIDEC dataset (In-domian)\cite{lalande2020emidec} and private dataset (out-of-domain) \cite{jathanna2023nottingham, shahnazprivatedata} in terms of Dice score. Where P: 57 real pathological cases. N’: (28$\times$2) synthetic pathological cases generated from 28 normal cases. N”: (28$\times$5) synthetic pathological cases generated from 28 normal cases. P’: (57$\times$1) synthetic pathological cases generated from 57 real pathological cases.}
\label{fig5}
\vspace*{-3mm}
\end{figure}

\noindent\textbf{Out-of-Domain Test:}
We further evaluated the performance of the segmentation models on the \emph{unseen}, \emph{large}, \emph{private} dataset \cite{jathanna2023nottingham, shahnazprivatedata}, consisting of  80 LGE-MRI provided by a local hospital. Fig.~\ref{fig5} compares the performance of different methods on the intra-domain EMIDEC (a) and the out-of-domain private test dataset (b) in terms of scar Dice scores for different subsets of synthetic data. It is clear that our method outperform the baseline on both the intra-domain but also the out-of-domain test sets. An interesting phenomenon we found is that when increasing the amount of synthetic data, the segmentation performance on out-of-domain data increased initially and then dropped. Moreover, our model based on joint training did not always produce better results on out-of-domain-data. The reason for worse performance on external data is mainly the significant volume differences. The scar volume is different in the EMIDEC dataset (refer to Table \ref{tab1:dataset}), and generating synthetic data with increased scar volume and distribution for the training of segmentation models may still lead to a data bias resulting in decreased performance on external data. In the future, we will consider to improve this with re-sampling techniques. 

\begin{figure}[!tbh]
\vspace*{-3mm}
\centering
\includegraphics[keepaspectratio=true, scale=0.9]{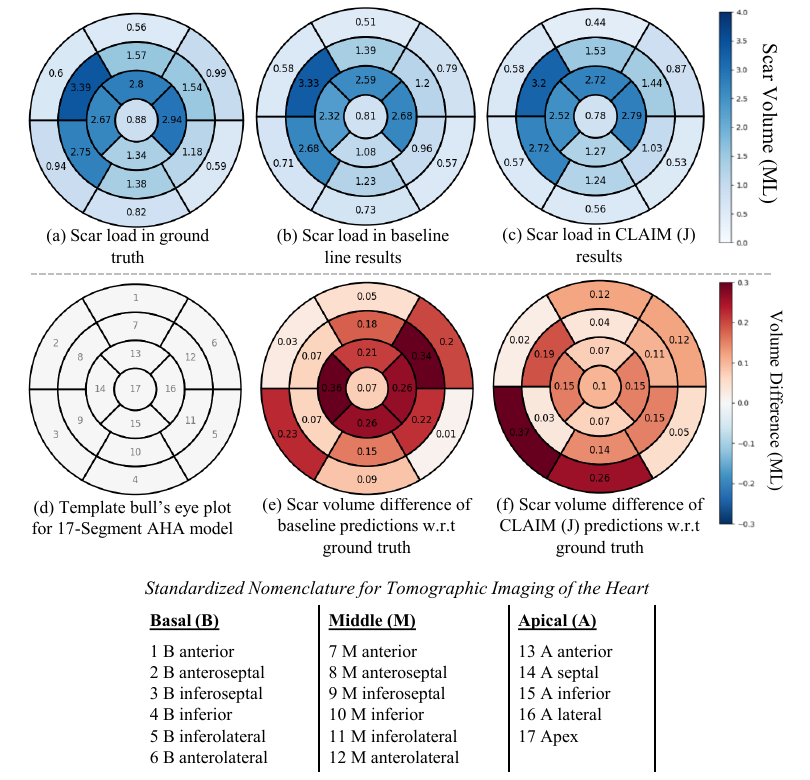}
\caption{Bull's eye plots for the evaluation of scar volume distribution with respect to AHA-17 segment model of left ventricular myocardium on EMIDEC test dataset.} \label{fig6}
\end{figure}

\noindent\textbf{Evaluation of Predicted Scar Volume Distribution:}
We also assessed the performance of our method with respect to the ground truth and the baseline method in terms of scar volume (ML) distribution using the AHA-17 segment framework. In Fig.~\ref{fig6}, we present bull's eye plots for standard 17 segment AHA model of left ventricular myocardium. Numbers on the plots quantitatively present the average volume of scars for each region of the myocardium. We have shown a template bull's eye plot showing region number and a table showing the name of the region in basal, middle and apical short axis slices according to standardized AHA nomenclature \cite{cerqueira2002standardized}.

The ground truth scar volume has been obtained from the EMIDEC test dataset while scar volume for baseline and CLAIM (J) have been obtained from their respective segmentation predictions. We have also shown volume difference with the ground truth in order to identify the difficult to segment regions. We observed that our method obtained better performance than baseline especially in the middle and apical regions (as indicated by the higher values). This analysis gives us insights into where our models succeed and fail and how can we leverage that to generate an optimized and diverse synthetic data distribution to better represent a wider population and varied pathology characteristics leading to the development of generalized scar synthesis and segmentation methods in the future.

\section{Conclusion}
We present CLAIM, a novel joint modeling framework that integrates diffusion-based image synthesis with myocardial scar segmentation to generate high-quality pathological LGE images. To guide the synthesis process, we proposed SMILE, a clinically informed mask generation method that leverages anatomical priors from the AHA 17-segment model to produce diverse and anatomically consistent scar patterns, varying in location, size, and texture. We further enhanced the diffusion training objective by incorporating a pathology segmentation loss, enabling more precise supervision from the synthesized scar regions. Our approach not only facilitates the generation of clinically plausible data but also improves the performance of state-of-the-art segmentation models such as nnUNet~\cite{isensee2021nnu} on both intra-domain and out-of-domain test sets, offering a scalable and effective solution in settings with limited annotated cardiac scar datasets.

\begin{credits}
\subsubsection{\ackname}
Farheen Ramzan acknowledges support from a PhD scholarship funded by the Commonwealth Scholarship Commission in the United Kingdom. Dr. Chen is supported by Royal Society (RGS/R2/242355). IT Services at The University of Sheffield provided High Performance Computing used in this study.
The authors have no competing interests to declare.
\end{credits}

\clearpage
\printbibliography
\end{document}